\documentclass[10pt,twocolumn,letterpaper]{article}

\usepackage{wacv}
\usepackage{times}
\usepackage{epsfig}
\usepackage{graphicx}
\usepackage{amsmath}
\usepackage{amssymb}
\usepackage{tabu} 
\usepackage{caption}
\usepackage{subfigure}
\usepackage{graphicx}

\wacvfinalcopy % *** Uncomment this line for the final submission

\def\wacvPaperID{****} % *** Enter the wacv Paper ID here

\ifwacvfinal\fi
\begin{document}

%%%%%%%%% TITLE
\title{Benchmark Visual Question Answer Models by using Focus Map}

% Authors at the same institution
%\author{First Author \hspace{2cm} Second Author \\
%Institution1\\
%{\tt\small firstauthor@i1.org}
%}
% Authors at different institutions
\author{Wenda Qiu \hspace{2cm} Yueyang Xianzang \hspace{2cm} Zhekai Zhang\\\\
Shanghai Jiaotong University\\
}

\maketitle
\ifwacvfinal\thispagestyle{empty}\fi

%%%%%%%%% ABSTRACT
\begin{abstract}
   Inferring and Executing Programs for Visual Reasoning proposes a model for visual reasoning that consists of a program generator and an execution engine to avoid end-to-end models. To show that the model actually learn which objects to focus on to answer the questions, the authors give a visualizations of the norm of the gradient of the sum of the predicted answer scores with respect to the final feature map. However, the authors does not evaluate the efficiency of focus map. This paper purposed a method for evaluating it. We generate several kinds of questions to test different keywords. We infer focus maps from the model by asking these questions and evaluate them by comparing with the segmentation graph. Furthermore, this method can be applied to any model if focus maps can be inferred from it. By evaluating focus map of different models on the CLEVR dataset, we will show that CLEVR-iep model has learned where to focus more than end-to-end models.

The main contribution of this paper is that we propose a focus-map-based evaluating method for visual reasoning.
\end{abstract}

%%%%%%%%% BODY TEXT
\section{Introduction}
Visual question answering(VQA) can be used in many applications. A VQA system takes an image with a natural language question about the image and give a natural-language answer as output\cite{VQA}. For example, it can be used to help visually-impaired poeple solving visual problems with talking mobile devices, or be a nearly real-time substitute for some human-powered services\cite{vizwiz}.

Existing methods for visual reasoning attempt to directly map inputs to outputs using black-box architectures without explicitly modeling the underlying reasoning processes. As a result, these black-box models often learn to exploit biases in the data rather than learning to perform visual reasoning. Inspired by module networks, the CLEVR-iep model gives a program generator that constructs an explicit representation of the reasoning process to be performed, and an execution engine that executes the resulting program to produce an answer. 

The author gives visualizations of the norm of the gradient of the sum of the predicted answer scores with respect to the final feature map to illustrate which object the model focus on when performing the reasoning steps for question answering. However, they do no evaluation on the effectiveness of the focus map, which leads the argument that CLEVR-iep has really learned concepts to be intuitive.
\begin{figure}[h]
    \begin{center}
         \includegraphics[width=0.5\textwidth]{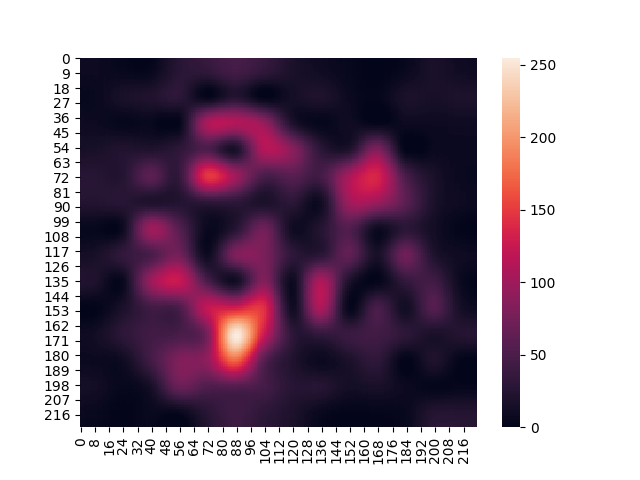}
         \caption{Visualizations of the norm of the gradient}
    \end{center}
\end{figure}

In this paper, we purposed a method to evaluate it. We generate several kinds of questions to test different keywords. These keywords include adjective directly describe objects such as 'blue' and 'cube' and adverbs which describe relative position such as 'behind' and 'on the left side'. At the same time we will get the answer object sets and do segmentation on origin pictures. Focus maps will be inferred from the model by asking these questions. We calculate focus value of focus maps by comparing with segmentation graph.  

\begin{figure}[h]
    \begin{center}
         \includegraphics[width=0.4\textwidth]{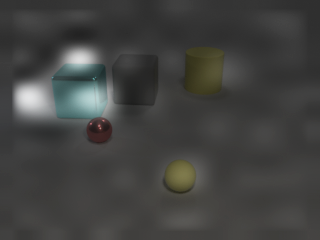}
         \caption{An illustration of focus map}
    \end{center}
\end{figure}

We do experiments on the CLEVR dataset. We test CLEVR-iep (18K prog. \& 700k prog.) model and CNN+LSTM model as baseline with different kinds of questions. The result shows that performance of CNN+LSTM model is nearly a random model and the other two models outperform it, which support the author's argument that black-box models often learn to exploit biases in the data rather than learning to perform visual reasoning.

\section{Related work}
\textbf{CLEVR} is a diagnostic dataset for VQA systems\cite{CLEVR}. The images in CLEVR are generated randomly using Blender based on a scene graph annotating the relations of objects. The questions and answers are generated by combinations of basic functional programs executed on an image’s scene graph. Additional tricks are used in avoiding ambiguity, ill-posed or degenerate questions and many other undesirable cases. CLEVR dataset is designed to analyze how a system learns sophisticated reasoning rather than biases in other VQA datasets. It provides us an approach to investigate abilities and limitations of a system.

\textbf{Gradient map} is suggested in Deep Inside Convolutional Networks: Visualising Image Classification Models and Saliency Maps\cite{simonyan2013deep}. The paper argues that the magnitude of the derivative indicates which pixels to be changed the least to affect the result most and it can be expected that such pixels correpsond to the object location in the image.

\textbf{CLEVR-iep} is a visual question answer model which explicitly incorporate compositional reasonging\cite{johnson2017inferring}. Specifically, it consists two parts: a program generator and an execution engine. The program generator reads the question and produces a program for answering question. The execution engine implements each function using a small neural module, and executes the resulting module network on the image to produce an answer. Using the CLEVR benchmark for visual reasoning, the author shows that CLEVR-iep model significantly outperforms strong baselines and generalizes better in a variety of settings.

\section{Method}
\subsection{Background}
The common evaluate method for VQA is calculate the accuracy. However, models can learn data biases but not reasoning to improve their accuracy. Inspired by Deep Inside Convolutional Networks: Visualising Image Classification Models and Saliency Maps\cite{simonyan2013deep}, which suggest that the gradient(which also called focus map in later part) of the result respect to the origin graph show the important objects' location, we propose an evaluate method based on the focus map to evaluate how much concept has VQA models actually learned. There are two challenges:how to evaluate the focus maps and how to use focus map to evaluate the VQA model. It can be observed that focus maps usually deviate from the centre of objects, which leads seperating accurate focus(which on the edges) and inaccurate focus(which outside the object but near the edges) be to more difficult. Also, it is easy to infer focus maps when evaluating adjectives describing objects directly are easy to evaluate but it is hard when evaluating adverb describing the relative relations between objects. We will show how our method solve these two challenges next. 
\subsection{Generating questions}
We use CLEVR to generate questions. The generating program needs a template of questions. For example, it can be "Is there a $\langle C \rangle$ object?" or "How many $\langle C \rangle$ objects are there?" or "Are there more $\langle C1 \rangle$ objects than $\langle C2 \rangle$ objects?" where $\langle C \rangle, \langle C1 \rangle, \langle C2 \rangle$ account for colors. These templates make model focus on objects by the same way that giving an adjective directly descride objects. To compare the model performance under different kinds of question, we will generate the 'exist' and 'how many' kind questions.

We will construct questions to evaluating adjectives and adverbs. To evaluate an adjective, suppose it is $\langle A_1 \rangle$, we will generate question like "Is there a $\langle A_1 \rangle$ objects?". If the model learns $\langle A_1 \rangle$ correctly(we will show how to measure 'correctly' in next parts), we will generate questions to evaluate adverbs based on the adjective $\langle A_1 \rangle$. Suppose the adverbs is $\langle R_1 \rangle$, the generated question will be "Is there a object $\langle R_1 \rangle$ of the $\langle A_1 \rangle$ objects?". We call it "one relationship question" since it has one relation word. 

If the model learns the adverb $\langle R_1 \rangle$ as well, we can add more adjectives. The generate question will be "Is there a $\langle A_2 \rangle$ object $\langle R_1 \rangle$ of the $\langle A_1 \rangle$ objects?", here $\langle A_2 \rangle$ is another adjective. According to the rules, we can generate "two,three,$\cdots$ relationship questions". Since most real questions only have at most one relation word and one relation word is enough to evaluate the model, we only generate questions with at most one relation word.

CLEVR also generates answers for questions automatically. It generates metadata of a picture which contains objects location, size, shap and color in it. The question generated will be converted to some suquences of filters and a final classifier(See Figure 3). The filtered metadata is exactly the ground truth foucsed objects.

\begin{figure}[h]
    \begin{center}
         \includegraphics[height=0.4\textheight]{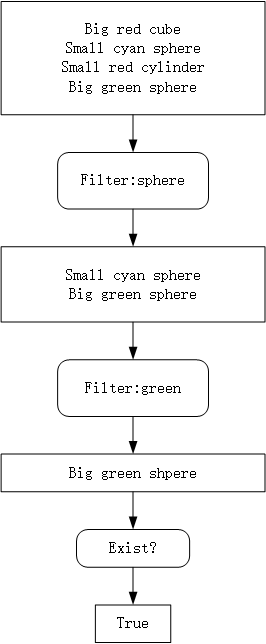}
         \caption{The filter sequence of question ``Is there a green sphere?" }
    \end{center}
\end{figure}

To get segmentation graphs, we render objects in pictures shadelessly and with different colors. Pixels have the same color of an object should be the pixels of it. We label pixels by its color's corresponding object to get segmentation.   

\subsection{Evaluate the model with Focus Map}
For each scene and a corresponding question, we can now get the Focus Map by calculating the gradient of the final output with respect to the feature map. We want to evaluate the model using the generated Focus Map and the ground truth focused objects constructed by the rule as mentioned above. 

Intuitively, we can sum up all the Focus Values (we call it Focus Sum later) inside an object. Since our scenes are generated by software, we can easily determine which object a point belongs to (see Figure 4) and therefore calculating the Focus Sum is possible. When the Focus Sum is larger than a threshold, we can assume the model focus on the object. By comparing it with the ground truth, we will get the correctness of the model.

\begin{figure}[h]
    \begin{center}
         \includegraphics[width=0.4\textwidth]{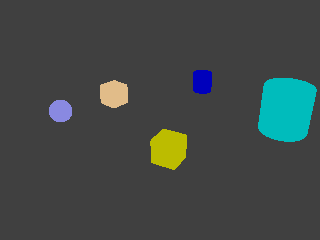}
         \caption{A segmentation graph}
    \end{center}
\end{figure}

Formally, we have a $W \times H$ Focus Map matrix $F$, of which the elements are the Focus Values of every points. The scene $S$ consists of many objects. Each object $x \in S$ is a set of points inside the object. We also have a ground truth objects set $T \subset S$.

We first normalize the Focus map to ensure
$$ \max_{0\le i<H, 0\le j<W} F(i,j) = 1 $$
For each object $x$, the normalized Focus Sum is 
$$ f(x) = \sum_{(i,j) \in x} F(i,j) / |x| $$
If there is a threshold $\theta$, the set of all the focused objects of the model is 
$$ \{x|f(x)\ge \theta\} $$
Actually, we can regard the problem as a 2-classification problem which classify all objects into “focused” and “unfocused”. By using AUC (Area Under the Curve) to evaluate the model, we don’t need to set up the threshold manually.

However, this method doesn’t work so well as expected in practice since the model will somethings focus on the edge of an object (see Figure 2), in which case the sum inside the object is relatively small. To solve the problem, we “blur” the objects using method like Gaussian Blur, the points outside an object is also taken into account but with a decay term $\eta < 1$.

Formally, for each point $(i,j)$ and each object $x$, we have decay term $\eta(i,j,x)$. Initially, all $(i,j) \in x$ has $\eta(i,j,x)=1$ and $\eta(i,j,x)=0$ otherwise. Then we apply Gaussian Blur to $\eta$ so that 
$$ \eta'(i,j,x) = \sum_{i',j'} \eta(i',j',x) \times \frac{1}{2\pi\sigma^2} e^{-((i-i')^2+(j-j')^2)/2\sigma^2} $$
And normalize $\eta$ to make sure
$$ \sum_{i,j} \eta'(i,j,x) = 1 \quad \forall x \in S $$
The normalize Focus Sum is
$$ f(x) = \sum_{i,j} \eta'(i,j,x) $$

\section{Experiment}
We evaluate models on a dataset with 250 images generated by CLEVR. 
\subsection{Generated questions}
There are four kinds of questions: either a ``exist" or ``count" question with zero or one relation word. 
The templates of these questions are:

``Is there a $Z$ $C$ $M$ $S$?"

``How many $Z$ $C$ $M$ $S$?"

``Is there a $Z_2$ $C_2$ $M_2$ $S_2$ [that is] $R$ the $Z$ $C$ $M$ $S$?"

``How many $Z_2$ $C_2$ $M_2$ $S_2$ [that is] $R$ the $Z$ $C$ $M$ $S$?".

Here, $\left( Z,C,M,S \right)$ accounts for $\left( size, color, material, shape \right)$. We generate two questions per each template and image.
\subsection{Models}
We evaluate CLEVR-iep (18K prog. \& 700k prog.) model and CNN+LSTM model.

\textbf{CNN+LSTM}: Images and questions are encoded using convolutional network (CNN) features and final LSTM hidden states, respectively. These features are concatenated and passed to a MLP that predicts an answer distribution.

This model is our baseline. 

\textbf{CLEVR-iep(18K prog.)}: Questions are passed to a program generator and images are passed to an execute engine. The program generator and the execute engine is pretrain jointly with 18K ground truth programs.

\textbf{CLEVR-iep(700K prog.)}: The program generator and the execute engine is pretrained jointly with 700K ground truth programs.
\subsection{Experiment result}
The result is shown in Table 1. 
\begin{table*}[h]
	\centering
		\begin{tabular}{|c|c|c|c|c|}
		\hline
		Models & exist(0 relation) & count(0 relation) & exist(1 relation) & count(1 relation)\\
		\hline
		\hline
		CNN+LSTM & 0.495 & 0.489 & 0.483 & 0.511\\
		CLEVR-iep(18K prog.) & 0.814 & 0.817 & 0.777 & 0.754\\
		CLEVR-iep(700K prog.) & 0.887 & 0.744 & 0.850 & 0.746\\
		\hline
		\end{tabular}
	
	\caption{The AUC of different models and quesitons.}
\end{table*}

From the table we can find that AUCs of the CNN+LSTM model are nearly 0.5, which is the AUC of a random model for a binary-classification problem.

The other two models out perform the CNN+LSTM model and CLEVR-iep(700K prog.) performs slightly better than CLEVR-iep(18K prog.) in exist(0 relation). We give the focus maps of three models for a "exist(0 relation)" problem(see Figure 5) to confirm our intuition. This may support the argument that end-to-end models often learn to exploit biases in the data rather than learning to perform visual reasoning.

However, CLEVR-iep(700K prog.) performs significantly worse than CLEVR-iep(18K prog.) while answering count(0 relation). This is not caused by the fault of our focus maps evaluation. Figure 6 are focus maps of two models for a same image and a same question. We can find that foci of CLEVR-iep(700K prog.) are more blurry. Worse performance may due to the fault of pretrain. 

\begin{figure*}[h]
	\centering
    \subfigure[The focus map of the CNN+LSTM model]{
         \includegraphics[width=0.3\textwidth]{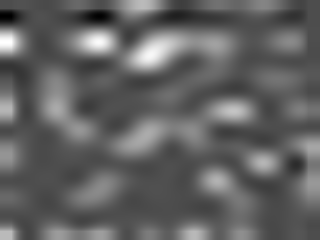}}
    \subfigure[The focus map of theCLEVR-iep(18K prog.) model]{
         \includegraphics[width=0.3\textwidth]{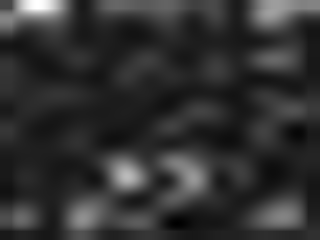}}
    \subfigure[The focus map of theCLEVR-iep(700K prog.) model]{
         \includegraphics[width=0.3\textwidth]{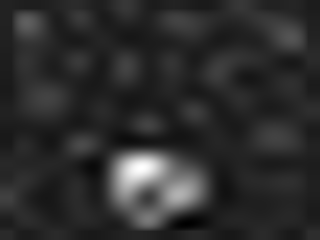}}
	\caption{Subfigure (a), (b), (c) are the focus maps of CNN+LSTM, CLEVR-iep(18K prog.), CLEVR-iep(700K prog.) for a same image and a same ``exist(0 relation)" problem. The CNN+LSTM model has nearly no focus, the CLEVR-iep(18K prog.) model has several blurry foci and the CLEVR-iep(700K prog.) model has one clear focus. }
\end{figure*}

\begin{figure*}[h]
	\centering
    \subfigure[The focus map of theCLEVR-iep(18K prog.) model]{
         \includegraphics[width=0.3\textwidth]{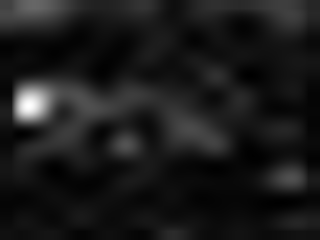}}
    \subfigure[The focus map of theCLEVR-iep(700K prog.) model]{
         \includegraphics[width=0.3\textwidth]{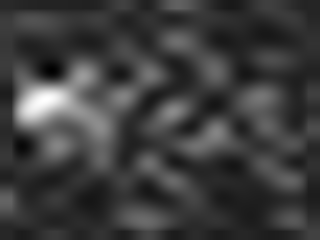}}
\end{figure*}

\section{Future Work}
	There is a significant decrease in AUCs between answering ``exist($x$ relation)" and ``count($x$ relation)"($x$ is either 0 or 1) problems. We did not generate these two kinds of questions with the same answer objects to find if ``count" is the reason of decrease. Our future work will generate these questions to find whether ``count" is the reason and why ``count" causes it.
\section{Conclussion}
This paper propose a focus-map-based evaluating method for visual reasoning. We use it to evaluate end-to-end models and non end-to-end models and find that non end-to-end models learn where to focus and end-to-end models do not learn it.
{\small
\bibliographystyle{ieee}
\bibliography{egbib}
}

\end{document}